\documentclass{article}
\usepackage{ijcai25}

\usepackage{times}
\usepackage{soul}
\usepackage{url}
\usepackage[hidelinks]{hyperref}
\usepackage[utf8]{inputenc}
\usepackage[small]{caption}
\usepackage{graphicx}
\usepackage{amsmath}
\usepackage{amsthm}
\usepackage{booktabs}
\usepackage{algorithm}
\usepackage{xcolor}
\usepackage{mathtools}
\usepackage[switch]{lineno}
\usepackage{makecell}
\usepackage{array}  
\usepackage{amssymb} 
\usepackage{algpseudocode}
\usepackage{amssymb}
\usepackage{placeins}

\title{Co-Sight: Enhancing LLM-Based Agents via Conflict-Aware Meta-Verification and Trustworthy Reasoning with Structured Facts}

\author{
Hongwei Zhang \and
Ji Lu \and
Shiqing Jiang\and
Chenxiang Zhu\and
Li Xie\and
Chen Zhong\and
Haoran Chen\and
Yurui Zhu\and
Yongsheng Du\and
Yanqin Gao\and
Lingjun Huang\and
Baoli Wang\and
Fang Tan,\And
Peng Zou \\
\affiliations
Zhongxing Telecom Equipment (ZTE), China\\
\emails
AIM@zte.com.cn
}

\begin{document}

\vspace{-2em}
\maketitle
\vspace{-3em}

\begin{abstract}

Long-horizon reasoning in LLM-based agents often fails not from generative weakness but from insufficient verification of intermediate reasoning.  Co-Sight addresses this challenge by turning reasoning into a falsifiable and auditable process through two complementary mechanisms: Conflict-Aware Meta-Verification (CAMV) and Trustworthy Reasoning with Structured Facts (TRSF).
CAMV reformulates verification as conflict identification and targeted falsification, allocating computation only to disagreement hotspots among expert agents rather than to full reasoning chains.  This bounds verification cost to the number of inconsistencies and improves efficiency and reliability.
TRSF continuously organizes, validates, and synchronizes evidence across agents through a structured facts module. By maintaining verified, traceable, and auditable knowledge, it ensures that all reasoning is grounded in consistent, source-verified information and supports transparent verification throughout the reasoning process. Together, TRSF and CAMV form a closed verification loop, where TRSF supplies structured facts and CAMV selectively falsifies or reinforces them, yielding transparent and trustworthy reasoning.
Empirically, Co-Sight achieves state-of-the-art accuracy on GAIA (84.4\%) and Humanity’s Last Exam (35.5\%) benchmarks, and strong results on Chinese-SimpleQA (93.8\%).  Ablation studies confirm that the synergy between structured factual grounding and conflict-aware verification drives these improvements.  Co-Sight thus offers a scalable paradigm for reliable long-horizon reasoning in LLM-based agents.\footnote{\textcolor{red}{Code available \href{https://github.com/ZTE-AICloud/Co-Sight/tree/cosight2.0_benchmarks}{\textbf{here}}}.}

\end{abstract}

\section{Introduction} \label{sec:intro}

Large Language Model (LLM)-based agents have made significant strides in solving complex, knowledge-intensive tasks, revolutionizing industries ranging from healthcare to education, finance, and beyond \cite{guo2024large,luo2025large,ferrag2025llm}. Their ability to process vast amounts of information, generate human-like text, and engage in reasoning across multiple domains has made them indispensable tools for various applications, including customer service, research assistance, and content generation. As these agents continue to evolve, their potential to support more advanced cognitive tasks—such as decision-making, problem-solving, and creativity—has far-reaching implications for both the global economy and technological innovation.

Despite rapid progress in long-horizon reasoning, recent large-scale agentic systems—including Deep Research Agents \cite{huang2025deep}, WebSailor \cite{li2025websailor}, WebResearcher \cite{qiao2025webresearcher}, and Dyna-Think \cite{yu2025dyna}—continue to exhibit systemic reliability bottlenecks once reasoning spans thousands of tokens or involves multiple external tools. One of the primary obstacles is the difficulty of performing scalable and effective verification, which currently applies to entire reasoning trajectories rather than focusing on key decision-making steps. This issue, when combined with the lack of structured context that combines retrieved evidence, assumptions, and intermediate computations, creates further barriers to achieving trustworthy attribution of reasoning. Specifically, these challenges manifest in three key limitations: (i) verification costs that scale with entire reasoning chains instead of focusing on the most crucial steps, (ii) the entanglement of context, where retrieved evidence, assumptions, and tool traces are not well-separated, leading to a lack of clarity, and (iii) the loss of provenance and consistency across different sources and tools, which compromises the reliability of final outputs. These limitations severely constrain the accuracy, interpretability, and cost-effectiveness of LLM-based agent systems in long-horizon reasoning tasks.  \par

Existing approaches for enhancing reliability remain fragmented. Semantic-entropy and self-verification approaches \cite{farquhar2024detecting,muhammed2025selfcheckagent} provide valuable signals for detecting hallucinations but can become costly or coarse when applied over long reasoning chains. Post-hoc entailment and embedding-based checks offer scalable alternatives yet often assess outputs only at the final-answer level, leaving local contradictions within intermediate reasoning steps unresolved. Meanwhile, recent planning-oriented scaffolds such as PlanGEN \cite{parmar2025plangen} and ALAS \cite{chang2025alas} introduce structured multi-agent or stateful planning mechanisms with limited integration of fine-grained verification. Their verification modules typically operate at the plan or task level rather than auditing specific inferential substeps. Similarly, Plan Verification \cite{hariharan2025plan} explores iterative critique and constraint checking, yet its trust attribution remains coarse-grained. As a result, current agent systems tend to over-audit low-impact steps while overlooking subtle conflicts that ultimately dominate outcome accuracy. \par

To address the limitations of current verification methods in agent-based reasoning, we propose Co-Sight, a conflict-aware and context-structured verification framework that optimizes computational cost by focusing on points of disagreement rather than the entire reasoning chain length. Co-Sight introduces a closed-loop architecture that explicitly decouples planning from auditing, treating the agent’s outputs as hypotheses to be falsified, rather than trusted answers, thereby enhancing verification efficiency and scalability. The core innovations of Co-Sight are the Conflict-Aware Meta-Verification (CAMV) and Trustworthy Reasoning with Structured Facts (TRSF) methods. CAMV optimizes the verification process by concentrating computational resources on minimal conflict sets identified by diverse expert agents. Instead of re-verifying entire reasoning chains, CAMV targets the key divergence points, making the process more computationally efficient and reliable. This approach significantly reduces the verification burden by focusing only on critical areas where conflicts arise. To provide the necessary foundation for CAMV, Co-Sight introduces the TRSF approach, which serves as a provenance-aware substrate for transparent and auditable reasoning. Instead of a static facts module, TRSF continuously organizes, validates, and synchronizes evidence across agents through a three-tier context-compression pipeline. This structure ensures that all reasoning operates on source-verified, traceable information, enabling consistent replay and independent verification. \textit{Our contributions are summarized as follows:}
\begin{itemize}
    \item CAMV: Verification is reformulated as conflict identification followed by targeted falsification, introducing constraint-based pruning, consensus anchoring, and conflict auditing. These mechanisms bound verification cost to the size of the disagreement set rather than full trajectory length, ensuring scalability and reliability.
    \item TRSF: The shared facts module continuously organizes, validates, and synchronizes evidence across agents. Through a three-tier context compression (tool, notes, and facts levels), it grounds reasoning in verified, traceable information, enabling transparent, auditable, and consistency-preserving inference.
\end{itemize}

Co-Sight~2.0 achieves state-of-the-art on General AI Assistants (GAIA) Test~\cite{mialon2023gaia} (84.4\%) and Humanity’s Last Exam (HLE)~\cite{phan2025humanity} (35.5\%), and performs strongly on Chinese-SimpleQA~\cite{he2024chinese} (93.8\%). Ablation studies indicate that the synergy between CAMV and TRSF underlies these gains, supporting the thesis that systematic auditing and context organization provide a more scalable path than further improvements in generation alone.


\section{Related Work}

\begin{figure*}[tb]
\centering
\includegraphics[scale=0.59]{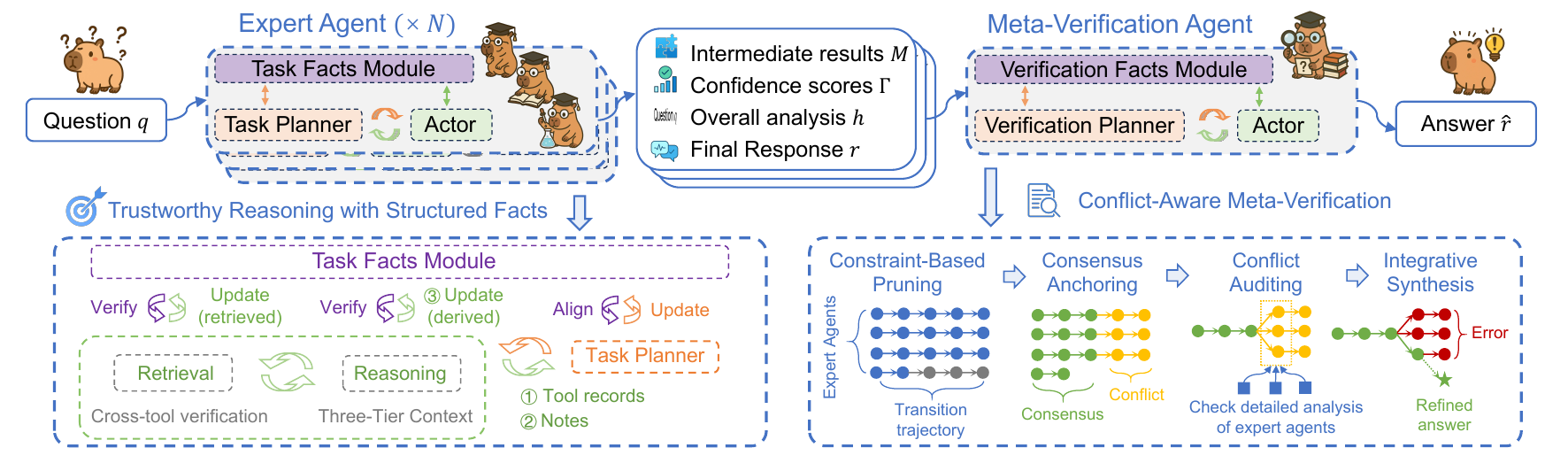}
\caption{\textbf{Overview of the Co-Sight Architecture Integrating CAMV and TRSF.} The framework forms a closed verification loop between multiple expert agents and a meta-verification agent. Each agent comprises a planner, actor, and a shared facts module that maintains source-verified knowledge. TRSF continuously records, summarizes, and validates evidence across agents, producing auditable factual anchors. The meta-verification agent applies CAMV, which localizes reasoning disagreements through constraint-based pruning, consensus anchoring, and conflict auditing. Only contentious nodes are re-verified, while validated anchors guide integrative synthesis of the final answer. This conflict-driven and fact-grounded interaction yields scalable, transparent, and trustworthy long-horizon reasoning.}
\label{fig:system}
\vspace{-7pt}
\end{figure*} 


\subsection{Agentic Reasoning and Planning}

Early reason-and-act scaffolds, typified by ReAct \cite{yao2023react}, interleave free-form reasoning with tool calls, enabling plans that adapt to observations. However, they often expand the search space indiscriminately and lack mechanisms to prioritize verification where it matters most. Graph-structured prompting (e.g., Graph-of-Thoughts, GoT) generalizes chain and tree methods by representing intermediate thoughts as vertices and dependencies as edges, improving search control and coordination across subtasks \cite{besta2024graph}. Surveys from 2024–2025 converge on two persistent gaps: (i) reliability for long-horizon planning and (ii) cost-aware control of verification within the loop \cite{plaat2024reasoning,verma2024brittle,deng2025simura}. Recent agent backbones (e.g., Plan-and-Act, ReflAct) incorporate explicit state-aware planning and iterative reflection, yet they continue to amortize verification broadly across many steps and trajectories \cite{erdogan2025plan,kim2025reflact}. \par

Co-Sight takes a complementary path. Rather than introducing another planning scaffold, it decouples verification from planning and concentrates computation on conflict hot spots that arise from divergence between conservative and radical views. In practice, it can be attached after any planner, providing tighter cost control while preserving planner choice.

\subsection{Self-Verification and Post-Hoc Checking}

Recent work examines post hoc verification for grounded generation. MiniCheck trains compact entailment models to verify sentence-level claims against grounding documents, matching GPT-4 performance at hundreds of times lower cost \cite{tang2024minicheck}. In a complementary line, CheckEmbed verifies open-ended outputs by comparing answer-level embeddings, offering a scalable, model-agnostic alternative to token-level checks \cite{besta2024checkembed}. Concurrent surveys clarify when generation-time control is preferable to post hoc correction and warn that “self-check everything” policies inflate cost and amplify early errors \cite{kamoi2024can,pan2024autocorrect}. \par

Co-Sight builds on these insights. Rather than blanket self-checking, CAMV focuses verification on disagreement sets revealed by agent diversity and source heterogeneity. Whereas MiniCheck and CheckEmbed verify every sentence or every answer, our meta-verifier first triages by conflict and spends its fixed budget only where the expected value of verification is high.

\subsection{Memory and Structured Context}

Retrieval-Augmented Generation (RAG) systems increasingly structure retrieved context to reduce noise and improve attribution. GraphRAG constructs a corpus-level knowledge graph that supports query-focused summarization and long-range aggregation \cite{edge2024local}, while LongRAG rebalances retrieval and reading by grouping documents into extended units that preserve global context \cite{jiang2024longrag}. Surveys of memory mechanisms for LLM agents catalogue persistent, summarized, and hierarchical memories, and highlight open problems in provenance tracking and cross-source conflict resolution \cite{zhang2025survey}. \par

Co-Sight mitigates inconsistencies and hallucinations by maintaining a shared facts module that ensures all information is consistently validated and grounded in reliable sources. A three-tier context compression mechanism, progressively extracting core information at each level to avoid the accumulation of irrelevant details in long contexts. This method prevents the loss of important information that can occur when attempting to summarize the entire context in a single layer.

\subsection{Multi-Agent System}

Debate-style ensembles can improve factual accuracy by eliciting argument among agents with a judge adjudicating outcomes, yet the benefits hinge on agent diversity and often erode under collusion or confirmation bias \cite{du2023improving,hegazy2024diversity,bandi2024adversarial,grigorian2025fail}. These frameworks typically optimize final answer accuracy, but they rarely reveal where agents substantively disagree or connect that signal to structured retrieval or memory. \par

Co-Sight reconceptualizes debate as a detection process in which conservative and radical agents jointly expose minimal conflict sets. Only those contentious spans are forwarded to a meta-verification module, transforming debate from an end-to-end solver into a cost-aware verification layer conditioned on source evidence and tool outputs.

\section{Methodology of Co-Sight} \label{sec:method}


Co-Sight comprises $N$ expert agents and a meta-verification agent. Given a user query $q$, the $n$-th expert produces a set of intermediate results $M_n$, a set of corresponding confidence scores $\Gamma_n$, an overall analysis $h_n$, and a response $r_n$. The meta-verification agent consumes
\begin{equation}
    \langle q, (M_1, \Gamma_1, h_1, r_1), \cdots, (M^{N}, \Gamma_{N}, h_{N}, r_{N}) \rangle,
\end{equation}
and synthesizes a final answer $\hat{r}$. The overall system workflow, including the interaction between expert agents and the meta-verification agent, is illustrated in Fig. \ref{fig:system}. Each agent consists of a planner, an actor, a toolkit, and a shared facts module. The planner decomposes the task into a concurrent Directed Acyclic Graph (DAG) $G = (S, E)$, where $S = \{s_1, s_2, \dots, s_k\}$ denotes the steps and $E$ captures the dependencies between them \cite{besta2024graph,yao2023tree}. The actor then executes these steps in a topological order, utilizing tools via the ReAct paradigm. After executing all steps, the planner extracts intermediate results $M=\{m_j\}$ with confidences $\Gamma=\{\gamma_j\}$, compiles an overall analysis $h$, and forms the candidate response $r$. In each agent, \textit{facts module} plays a critical role in ensuring the reliability of information at both the information retrieval and reasoning stages. \par

The core innovation of this framework lies in the CAMV mechanism within the meta-verification agent and the TRSF algorithm in the expert agent.

\subsection{Conflict-Aware Meta-Verification} \label{sec:meta-verification}

\subsubsection{Concept and Formalization}
Meta-verification refers to an agent’s capacity to retroactively evaluate its own reasoning and outputs by transforming conclusions into premises that can be tested against constraints or evidence \cite{shinn2023reflexion,2023selfcheckgpt}. This mechanism rests on two complementary principles. The first, self-consistency, posits that internally coherent reasoning is statistically more reliable and thus mitigates hallucination \cite{liang2024internal}. The second, verification-over-generation, holds that checking is typically lower-variance and lower-cost than producing in complex tasks \cite{gero2023self}. Within CAMV, these principles materialize through the auditing of disagreements and the promotion of recurrent intermediate results to anchoring premises. \par

Let $\{c_n\}_{n=1}^N$ be the set of candidate solutions of $N$ expert agents. Each candidate induces a set of reasoning steps indexed by $S$. For candidate $n$ and step $s\in S$, denote the intermediate result by $m_n(s)$, its local analysis by $h(m_n(s))$, and a calibrated confidence by $\sigma(m_n(s))\in[0,1]$. Let $M=\{m_n(s): n\in[N],\, s\in S\}$ collect all intermediates, and $\Gamma=\{\sigma(m):m\in M\}$ their confidence. Each candidate outputs a final answer $r_n$; the system produces a synthesized result $\hat{r}$. Domain knowledge and feasibility constraints, including schemas, units, invariants, and consistency rules, are represented by $\mathcal{K}$, against which all results are evaluated.

\begin{algorithm}[t]
\caption{CAMV: Conflict-Aware Meta-Verification}
\label{alg:cfmv}
\begin{algorithmic}[1]\small
\Require Query $q$; experts $\mathcal{E}$; tools $\mathcal{I}$; facts $\mathcal{F}$; gate ${\rm Gate}(\cdot)$; threshold $\theta$; budget $B_{\max}$; operators $\mathcal{V}$
\Ensure Final answer $\hat{r}$, confidence $s$, anchors $A_{\theta}$, conflicts $S_c$, screening log $\mathcal{E}_v$

\State $A_{\theta}\!\gets\!\emptyset,\ S_c\!\gets\!\emptyset,\ \mathcal{E}_v\!\gets\!\emptyset,\ \mathcal{T}_r\!\gets\!\emptyset$

\Statex \textbf{Stage 1: Constraint-Based Pruning}
\For{$e \in \mathcal{E}$}\For{$T' \in \textsc{SampleTraces}(e,q)$}
  \State $(\tilde{T}, s_g)\!\gets\! {\rm Gate}(T',q,\mathcal{F})$
  \If{$\tilde{T}\neq\emptyset$} \State $\mathcal{T}_r \!\gets\! \mathcal{T}_r \cup \{(\tilde{T}, s_g)\}$
  \Else\ \ \ \ \ \ \ \ \ \ \ \ \ \ \ \ \ \ \ \ \ \ \ \State $\mathcal{E}_v \!\gets\! \mathcal{E}_v \cup \{(\text{reject}, T')\}$
  \EndIf
\EndFor\EndFor

\Statex \textbf{Stage 2: Consensus Anchoring}
\State $\mathcal{U}\!\gets\!\textsc{Statements}(\mathcal{T}_r)$
\State $A_{\theta}\!\gets\!\{\,u\!\in\!\mathcal{U}\ :\ {\rm supp}(u)\!\ge\!\theta\,\}$

\Statex \textbf{Stage 3: Conflict Auditing}
\State $S_c \!\gets\! \textsc{Conflicts}(\mathcal{T}_r) \setminus A_{\theta}$ \Comment{steps where experts disagree}
\State $b\!\gets\!0$
\For{$u \in \textsc{Rank}(S_c)$}
  \If{$b \ge B_{\max}$} \textbf{break} \EndIf
  \State $v \!\gets\! \textsc{Verify}(u,A_{\theta},\mathcal{I},\mathcal{F},\mathcal{V})$ \Comment{support / refute}
  \If{$v=\text{support}$} \State $A_{\theta} \!\gets\! A_{\theta}\cup\{u\}$
  \ElsIf{$v=\text{refute}$} \State $S_c \!\gets\! S_c\cup\{u\}$
  \EndIf
  \State $b \!\gets\! b+1$
\EndFor

\Statex \textbf{Stage 4: Integrative Synthesis}
\State $(\hat{r},s)\!\gets\!\textsc{Synthesize}(q,A_{\theta},S_c,M,\Gamma,H,\mathcal{I},\mathcal{F})$
\Comment{see Eq.~(6)}
\State \Return $\hat{r}$

\end{algorithmic}
\end{algorithm}


\subsubsection{Pipeline}

The naive item-by-item checking is replaced with a structured four-stage pipeline tailored to complex reasoning tasks such as HLE and GAIA. The detailed algorithm is represented in Algorithm \ref{alg:cfmv}.

\textit{Constraint-Based Pruning}. Instead of discarding complete reasoning traces, the procedure excises only those intermediates that violate domain constraints and prunes their dependent derivations, while retaining all preceding valid segments. Formally, let $\mathcal{D} = \{\, r \mid r \models \mathcal{K} \,\}$ denote the set of results consistent with the constraint set $\mathcal{K}$. Candidate answers are filtered according to

\begin{equation}
R=\{\, r_i \in \{r_1,\dots,r_N\}\mid r_i\in \mathcal{D}\,\}.
\label{eq:retained}
\end{equation}
If $R=\emptyset$, the system triggers Elimination-by-Aspects (EBA) backtracking, which iteratively locates violated constraints, removes dependent intermediates, and repairs only the affected sub-chains—thereby avoiding a full re-parse.

\textit{Consensus Anchoring}. Recurring intermediates are promoted to anchors to reduce verification load. With threshold $\theta\in\{2,\cdots,N\}$, define
\begin{equation} 
\mathcal{A}_{\theta}
= \{\, (s, v) \in \mathcal{U} \mid 
|\{\, i : m_i(s) = v \,\}| \ge \theta \,\},
\label{eq:consensus}
\end{equation}  
where $\mathcal{U} = \mathrm{STATEMENTS}(T_r)$ and elements of $\mathcal{A}_{\theta}$ are treated as verified premises for downstream checks.

\textit{Conflict Auditing}. Verification is confined to nodes where candidate reasoning traces diverge:
\begin{equation}
S_{\mathrm{c}}
=\{\, s\in S \mid \exists i\ne j \quad {\rm s.t.} \quad \ m_i(s)\neq m_j(s) \,\}.
\label{eq:conflict}
\end{equation} \par

The auditor allocates its budget to $S_c$, applying EBA-style falsification tests (unit checks, constraint satisfaction, cross-tool re-execution). This concentrates effort where it affects outcomes and bounds cost by $|S_c|$ rather than the full chain length $|S|$.

\textit{Integrative Synthesis}. When every candidate exhibits faults at the audited nodes, CAMV reconstructs a coherent reasoning trace by extracting valid micro-inferences from $M$, recombining them under the domain constraints $\mathcal{K}$ and the facts set $\mathcal{F}$, and consolidating the resulting evidence into a unified answer. A scoring function $\mathrm{Score}(\cdot)$ integrates anchor support, resolved-conflict evidence, and calibrated confidence estimates. The final response is obtained as

\begin{equation}
\hat{r}
= \arg\max_{r \in \mathcal{D}} \;
\mathrm{Score}\!\left(
r;\,
\mathcal{A}_{\theta},\, S_c,\, M,\, \Gamma,\, H
\right),
\label{eq:score}
\end{equation}
with $H=\{h_n\}_{n=1}^N$. This converts imperfect candidates into useful signals and yields a traceable final decision.

\subsubsection{Conservative–Radical Ensemble}

To ensure that conflicts serve as informative diagnostic signals rather than noise, experts are instantiated from a single base LLM under diversified temperature settings. Low-temperature (conservative) experts emphasize stability and precision, establishing high-confidence baselines and reliable anchors for verification. In contrast, higher-temperature (radical) experts expand coverage over low-probability yet potentially correct reasoning trajectories, thereby enriching the diagnostic content of the disagreement set $S_c$. \par

In practical configurations, a small fixed portion of the computational budget is reserved for conservative agents to prevent dominance while maintaining persistent disagreement. Within the CAMV framework, conservative predictions reinforce $\mathcal{A}_{\theta}$ by consolidating verified anchors, whereas radical outputs enlarge $S_c$ to expose divergent reasoning chains. The verifier subsequently adjudicates only these contested points, balancing efficiency with epistemic diversity.


\subsection{Trustworthy Reasoning with Structured Facts} \label{sec:agent-design}

\begin{figure}[tb]
\centering
\includegraphics[scale=0.55]{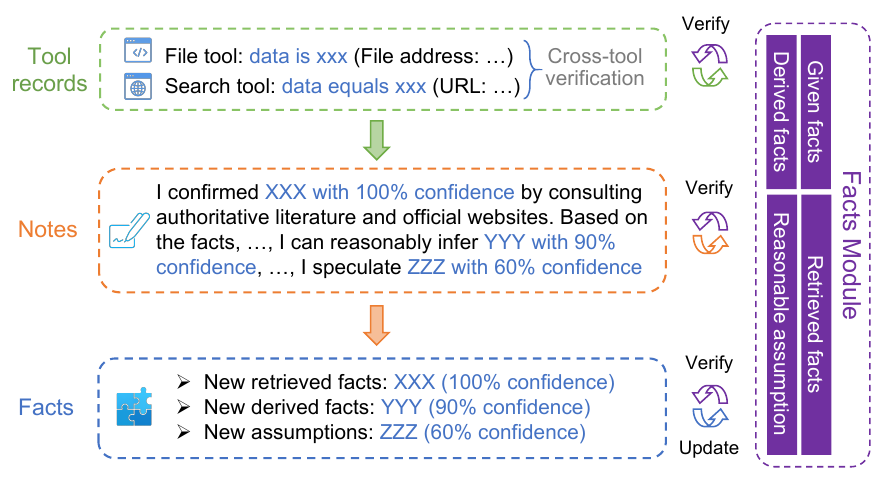}
\caption{\textbf{Illustration of TRSF}. Co-Sight gradually extracts the underlying tool records into new facts, and uses the facts module to verify inconsistencies and check for hallucinations.}
\label{fig:workflow_agent}
\vspace{-5pt}
\end{figure}

The effectiveness of CAMV hinges on the reliability of the evidence it relies on. TRSF provides this evidential substrate by maintaining a provenance-aware facts module that organizes, validates, and synchronizes knowledge across agents (as illustrated in Fig. \ref{fig:workflow_agent}). Together, TRSF and CAMV form a closed loop: TRSF supplies structured, auditable facts, and CAMV selectively falsifies or reinforces them. \par

\subsubsection{Structure and Role of the Facts Module}
The shared facts module organizes information into four categories: given facts, retrieved facts, derived facts, and assumptions. This module is continually updated and anchors long-horizon reasoning in verifiable information rather than transient model outputs, reducing hallucination and inconsistency.

\subsubsection{Information Retrieval}

In the information retrieval phase, the credibility of the gathered data is ensured through a rigorous cross-verification process conducted at the data source level. Specifically, provenance tracking mechanisms capture metadata, including URLs and download logs for each item retrieved. This process provides a foundation for subsequent reasoning and the updating of facts.

\subsubsection{Facts-Based Reasoning}
Throughout the reasoning process, all new intermediate data will be compared with the facts stored within the facts module. In the event of discrepancies, additional verification steps are performed to ensure that the data remains consistent and logically coherent. This iterative comparison guarantees that the facts within the module are both reliable and accurate, ensuring that all reasoning is rooted in verifiable, logically sound information. \par

To continuously extract and refine factual knowledge throughout long-horizon reasoning, Co-Sight employs a three-tier context compression mechanism. Unlike traditional approaches that either retain excessive noise or discard useful detail, this mechanism progressively distills essential information from raw reasoning traces, yielding concise yet verifiable factual representations.

\begin{itemize}
    \item Tool Level: Essential but minimal metadata is recorded, such as the identity of the tool used, its parameters, and the outcomes achieved.
    \item Notes Level: The trajectory is summarized into concise annotations that reflect state and credibility judgments. 
    \item Facts Level: Only stable and verified knowledge is incorporated into the shared facts module for reuse in subsequent reasoning. 
\end{itemize}

This hierarchical compression turns traces into structured knowledge, enabling efficient transfer and lowering verification effort by letting CAMV focus on trusted anchors.

\begin{figure*}[tb]
\centering
\includegraphics[scale=0.46]{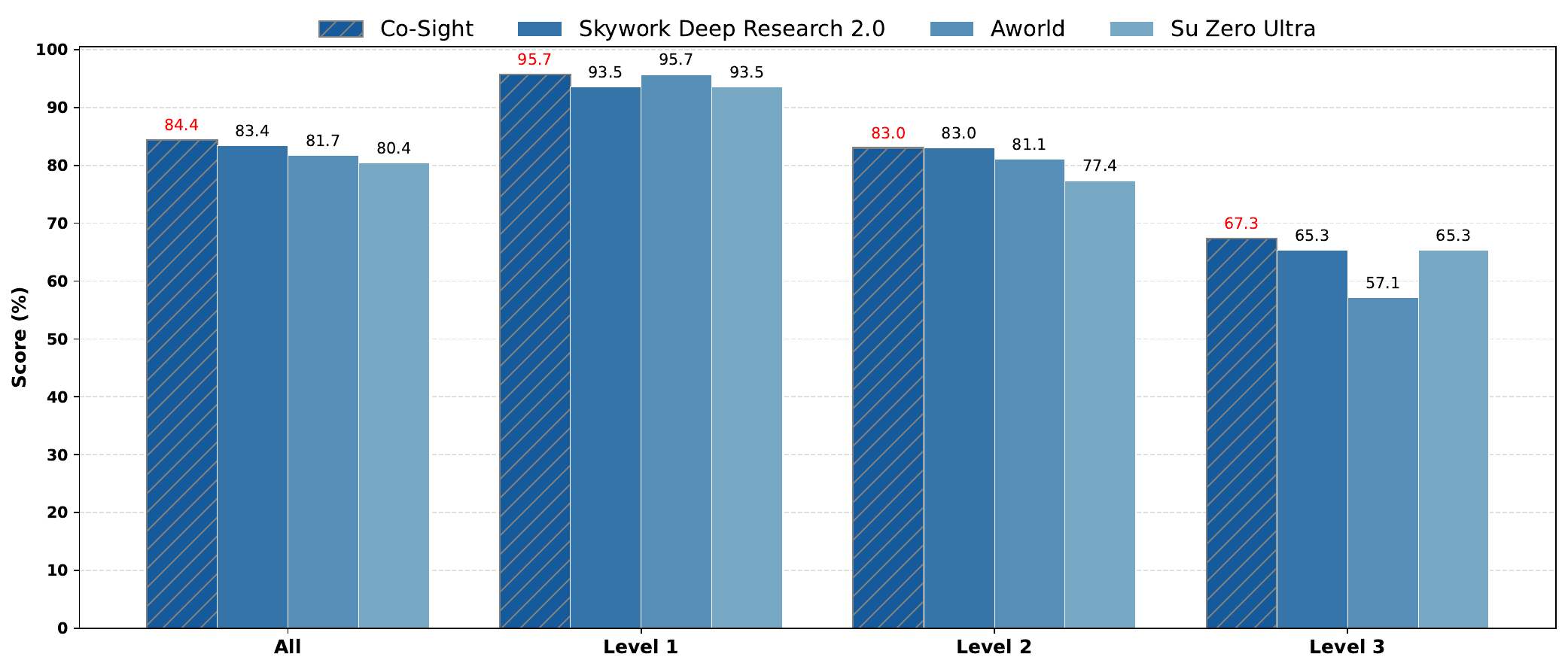}
\caption{Performance comparison on the GAIA test benchmark.}
\label{fig:GAIA_results}
\vspace{-6pt}
\end{figure*}

\section{Experimental Evaluation} \label{sec:results}

\subsection{Datasets}

Evaluation of Co-Sight is conducted on three representative benchmarks:

\begin{itemize}
    \item GAIA \cite{mialon2023gaia}: 300 real-world questions spanning three difficulty levels, designed to assess tool-augmented reasoning in general AI assistants.
    \item HLE \cite{phan2025humanity}: a closed-ended, interdisciplinary suite covering 100+ fields that stresses advanced reasoning and multimodal understanding at the frontier of human knowledge.
    \item Chinese-SimpleQA \cite{he2024chinese}: 3,000 factual questions across six domains (e.g., culture, natural sciences, engineering), emphasizing Chinese factuality with primarily single- or double-hop retrieval, thereby probing exploration reliability.
\end{itemize}

\subsection{Results on GAIA}

Performance of Co-Sight is assessed on the GAIA test benchmark\footnote{The official GAIA leaderboard is available at \url{https://huggingface.co/spaces/gaia-benchmark/leaderboard}.} in comparison with leading agentic systems, including Skywork Deep Research v2 \cite{zhang2025agentorchestra}, AWorld \cite{yu2025aworld}, and Su Zero Ultra. Fig. \ref{fig:GAIA_results} presents the comparative accuracy across difficulty levels. Co-Sight attains the highest overall score of 84.4\%, exceeding the next best system by 1.0\% and sustaining clear superiority across all task categories.

\subsubsection{Performance on Easy Tasks} 
For questions that require factual retrieval and moderate reasoning (Level 1 and Level 2), Co-Sight’s advantage arises primarily from two mechanisms. The first is the domain-gated credibility ranking module, which filters irrelevant or low-reliability information before reasoning begins, ensuring that downstream reasoning chains rely on verified inputs. The second is the multi-layer context compression mechanism, which organizes retrieved evidence hierarchically into tool-level metadata, concise notes, and validated facts. This structure enables cross-domain inference and prevents redundant analysis. As a result, the system constructs a coherent global reasoning outline before performing detailed steps, thereby reducing computational cost and minimizing error propagation. Empirically, these mechanisms explain the strong Level 1–2 results (95.7\% and 83.0\%), as more computation is directed toward verifying key premises instead of exploring unproductive alternatives. \par

\subsubsection{Performance on Difficult Tasks} 
For Level 3 items involving multi-hop retrieval and long-range logical composition, Co-Sight achieves 67.3\% accuracy, outperforming Skywork Deep Research v2 (65.3\%) and AWorld (57.1\%). The improvement results from the fact-driven and conflict-aware verification strategy. Instead of re-examining entire reasoning chains, the meta-verifier allocates computation only to steps where experts disagree, which reduces complexity from the total chain length to the number of contentious nodes. Furthermore, the fact-driven verification loop mitigates hallucinations by cross-checking retrieved facts against domain constraints and external tools such as code execution and literature lookup. This process stabilizes multi-stage retrieval accuracy and preserves coherence in extended reasoning contexts. \par

The GAIA results demonstrate that Co-Sight achieves state-of-the-art performance by strategically allocating verification resources. It applies pruning and anchoring for simple tasks, structured context integration for moderately complex reasoning, and conflict-aware auditing for long-horizon compositions. The observed accuracy margins are consistent with the system’s design principle of auditing rather than regenerating knowledge.

\subsection{Results on HLE}

\begin{figure}[tb]
\centering
\includegraphics[scale=0.33]{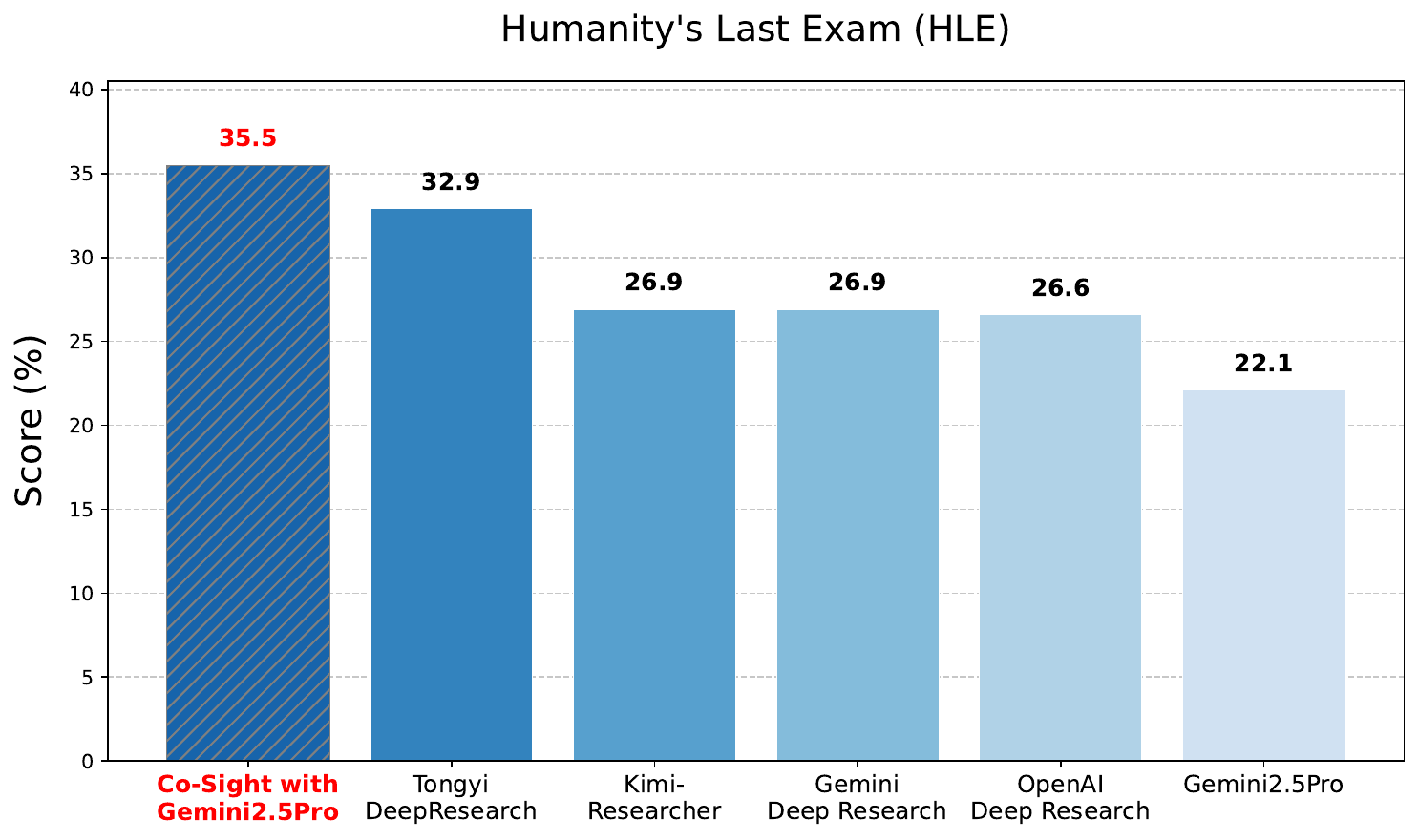}
\caption{Performance Comparison on the HLE Benchmark.}
\label{fig:HLE_results}
\vspace{-6pt}
\end{figure}

On the HLE benchmark, Co-Sight was evaluated against leading agentic frameworks such as Tongyi DeepResearch, Kimi-Researcher \cite{moonshot2025kimi-researcher}, Gemini Deep Research \cite{google2025gemini-deep-research}, and OpenAI Deep Research. To disentangle model- and framework-level contributions, the Gemini 2.5 Pro backbone used in Co-Sight was also tested as a control. As shown in Fig. \ref{fig:HLE_results}, Co-Sight achieves the highest score of 35.5\%, outperforming Tongyi DeepResearch (32.9\%) and substantially exceeding the Gemini 2.5 Pro baseline (22.1\%). \par

The HLE benchmark comprises challenging interdisciplinary tasks that require multistep reasoning and the integration of multimodal evidence. Co-Sight improves performance through two pillars. First, CAMV audits only the divergent nodes across expert reasoning traces rather than entire paths, following principles of self-consistency and verification complexity. Combined with coarse-grained reverse elimination and elimination by aspects (EBA), it isolates violated constraints and efficiently repairs the affected subchains. Second, the TRSF approach manages context in three phases: local extraction, cross-task integration, and global storage, thereby building a factual repository of validated intermediates. A multi-source trustworthiness scheme then cross-checks results across tools and sources, including code execution and literature retrieval, improving both conflict-detection accuracy and verification reliability. Collectively, these components convert reasoning from opaque generation into a transparent, auditable process that directly supports CAMV’s requirement for reliable evidence. \par

Together, these mechanisms form a closed loop of factual support, trustworthy input, and precise verification. This process satisfies HLE’s stringent requirements for high-order reasoning. The observed gain of 2.6\% over the strongest competitor and 13.4\% over the backbone model validates Co-Sight’s central premise that strategically allocated verification and conflict-driven auditing can outperform purely generative reasoning in complex environments.

\subsection{Ablations on Chinese-SimpleQA}

The Chinese-SimpleQA benchmark is employed to analyze the sources of performance gains in Co-Sight.

\subsubsection{Expert Ensemble Size}

To investigate scalability, the analysis begins by examining how accuracy varies with the number of expert agents $N$. All methods share identical sampling configurations (same base LLM and temperature).The oracle-style pass@N reports success if any candidate among N matches the ground truth \cite{chen2021evaluating}. Simple Verification (SV) uses the same input as CAMV but skips its four-stage verification pipeline, spontaneously generating the answer. Majority Voting (MV) chooses the most frequent prediction \cite{wang2022self}. \par

As shown in Table \ref{tab:expert_agents_performance}, the score of CAMV rises from 88.3\% to 91.2\% when $N=1\to 2$, surpassing pass@$N$ under this small-ensemble setting. This improvement indicates that conflict-aware auditing effectively recovers and recombines partial micro-inferences that single trajectories alone fail to consolidate. For larger ensembles ($N\geq 3$), pass@$N$ outperforms CAMV, which is expected: pass@$N$ is rewarded for any correct sample, whereas CAMV must allocate a finite audit budget $B_{\rm max}$ across an expanding set of disagreements. As $N$ grows, both the anchor pool $\mathcal{A}_{\theta}$ and the conflict set increase, reducing the fraction of conflicts that can be thoroughly audited within fixed compute limits. Hence, Co-Sight’s advantage at $N\leq 2$ demonstrates that selective, conflict-centric verification remains compute-efficient in practical ensemble regimes.

\begin{table}[t]
  \centering
  \caption{\textbf{Score (\%) on Chinese-SimpleQA across expert agent ensemble sizes ($N$).} CAMV outperforms pass@$N$ for smaller $N$.}
  \label{tab:expert_agents_performance}
  \begin{tabular}{ccccc}
    \hline
    $N$ & $1$ & $2$ & $3$ & $4$ \\
    \hline
    CAMV (ours) & \textbf{88.3}$\uparrow$ & \textbf{91.2}$\uparrow$ & 93.1 & 93.8 \\
    Oracle-style pass@$N$ & 85.0 & 90.0 & \textbf{93.3}$\uparrow$ & \textbf{94.5}$\uparrow$ \\
    SV & 86.7 & 88.9 & 90.2 & 90.6 \\
    Majority Voting & 85.0 & 85.0 & 86.7 & 87.8 \\
    \hline
  \end{tabular}
  \vspace{-5pt}
\end{table}

\subsubsection{Component Ablations}

The analysis next isolates the contributions of three core modules: the CAMV mechanism, the SV baseline, and the TRSF method. Since CAMV inherently subsumes the functionality of SV, enabling CAMV implicitly activates SV. \par

As listed in Table \ref{tab:method_accuracy}, the full Co-Sight (CAMV + TRSF) achieves the highest accuracy of 91.2\%, confirming the synergy between fine-grained verification and hierarchical context organization. Replacing the CAMV pipeline with a single-step verifier yields a 3.0\% drop, underscoring that focusing compute on divergent nodes rather than full-path re-checks is more efficient. Removing TRSF while retaining CAMV causes an additional 2.4\% decrease, as unstructured dialogue accumulates contextual noise and prolongs planner–actor exchanges. In contrast, configurations retaining only SV or TRSF degrade further (–5.5 and –6.2\%, respectively). These trends reveal a strong interdependence: verification relies on structured, low-noise evidence, while structured context realizes its full potential only under an active verifier. Collectively, integrating both CAMV and TRSF produces an 8.6\% gain over the baseline, validating the design principle of focused verification + structured reasoning as the foundation for scalable factual intelligence.


\begin{table}[t]
\centering
\caption{
\textbf{Component Ablations on Chinese-SimpleQA ($N=2$ experts).}
Adding CAMV and TRSF jointly yields an overall gain of +8.6\% over the baseline.
}
\resizebox{\linewidth}{!}{
\begin{tabular}{ccccccc}
\toprule
\textbf{Configuration} &
\textbf{TRSF} &
\textbf{SV} &
\textbf{CAMV} &
\textbf{Score (\%)} &
\textbf{$\Delta$} \\
\midrule
Baseline       & --             & --    & --    & 82.6 & -- \\
SV           & -- &   \checkmark   & --    & 85.7 & +3.1 \\
TRSF              & \checkmark   & --          & --    & 85.0 & +2.4 \\
CAMV                       & --           & \checkmark    & \checkmark     & 88.8 & +6.2 \\
SV + TRSF         & \checkmark    & \checkmark     & --     & 87.7 & +5.1 \\
\textbf{Co-Sight}  & \checkmark & \checkmark  & \checkmark & \textbf{91.2} & \textbf{+8.6} \\
\bottomrule
\end{tabular}
}
\vspace{-5pt}
\label{tab:method_accuracy}
\end{table}

\section{Limitations and Broader Impact}

\subsection{Limitations}
Co-Sight relies on the precise detection and alignment of points of disagreement across reasoning traces. If experts produce incomplete plans, the resulting conflict set can miss salient errors, which narrows audit coverage. Moreover, the multimodal pipeline remains bounded by the accuracy of current vision and parsing modules. Finally, although the method yields consistent gains on benchmarks such as GAIA and HLE, its robustness in real-world, safety-critical settings has yet to be established.

\subsection{Broader Impact}
By reallocating computation from exhaustive rechecking of full paths to auditing focused on conflicts, and by exposing anchors, constraints, and provenance at the tool level, Co-Sight promotes transparency and accountability in agentic systems. It can improve reliability and auditability across complex reasoning pipelines, with benefits for both research and practice. Expected benefits include safer assistants that use retrieval augmentation, clearer error surfaces for human reviewers, and better trade-offs between cost and quality for reasoning over long horizons in education, scientific curation, and enterprise analytics. Overall, Co-Sight advances accountable reasoning audits, but it is not a substitute for expert oversight or domain-specific governance.

\section{Conclusion} \label{sec:conclusion}

This paper introduces Co-Sight, a closed-loop cognitive architecture that shifts LLM agents from generating answers to auditing reasoning. Using the proposed CAMV algorithm, the system verifies only critical divergence points rather than rechecking entire reasoning chains, improving efficiency and reliability. With TRSF mechanism, Co-Sight supports transparent, auditable, long-horizon reasoning. \par

Co-Sight achieves state-of-the-art results on GAIA test and HLE, with accuracies of 84.4\% and 35.5\%, respectively, and maintains strong performance on Chinese-SimpleQA (93.8\%). These findings indicate that conflict-driven verification and structured evidence organization can outperform purely generative reasoning under comparable computational budgets. Future work will investigate adaptive verification budgets, stronger multimodal verifiers, and deployment in safety-critical domains to further improve the robustness and accountability of autonomous reasoning systems.

\newpage

\bibliographystyle{named}
\bibliography{refer.bib}

\begin{thebibliography}{}

\bibitem[\protect\citeauthoryear{AI}{2025a}]{google2025gemini-deep-research}
Google AI.
\newblock Gemini deep research: Automated research assistant.
\newblock Technical report, 2025.

\bibitem[\protect\citeauthoryear{AI}{2025b}]{moonshot2025kimi-researcher}
Moonshot AI.
\newblock Kimi-researcher: End-to-end {RL} training for emerging agentic capabilities.
\newblock Technical report, Moonshot AI, 2025.

\bibitem[\protect\citeauthoryear{Bandi and Harrasse}{2024}]{bandi2024adversarial}
Chaithanya Bandi and Abir Harrasse.
\newblock Adversarial multi-agent evaluation of large language models through iterative debates.
\newblock {\em arXiv preprint arXiv:2410.04663}, 2024.

\bibitem[\protect\citeauthoryear{Besta \bgroup \em et al.\egroup }{2024a}]{besta2024graph}
Maciej Besta, Nils Blach, Ales Kubicek, Robert Gerstenberger, Michal Podstawski, Lukas Gianinazzi, Joanna Gajda, Tomasz Lehmann, Hubert Niewiadomski, Piotr Nyczyk, et~al.
\newblock Graph of thoughts: Solving elaborate problems with large language models.
\newblock In {\em AAAI conference on artificial intelligence (AAAI)}, volume~38, pages 17682--17690, 2024.

\bibitem[\protect\citeauthoryear{Besta \bgroup \em et al.\egroup }{2024b}]{besta2024checkembed}
Maciej Besta, Lorenzo Paleari, Marcin Copik, Robert Gerstenberger, Ales Kubicek, Piotr Nyczyk, Patrick Iff, Eric Schreiber, Tanja Srindran, Tomasz Lehmann, et~al.
\newblock {CHECKEMBED}: Effective verification of {LLM} solutions to open-ended tasks.
\newblock {\em arXiv preprint arXiv:2406.02524}, 2024.

\bibitem[\protect\citeauthoryear{Cemri \bgroup \em et al.\egroup }{2025}]{grigorian2025fail}
Mert Cemri, Melissa~Z Pan, Shuyi Yang, Lakshya~A Agrawal, Bhavya Chopra, Rishabh Tiwari, Kurt Keutzer, Aditya Parameswaran, Dan Klein, Kannan Ramchandran, et~al.
\newblock Why do multi-agent {LLM} systems fail?
\newblock {\em arXiv preprint arXiv:2503.13657}, 2025.

\bibitem[\protect\citeauthoryear{Chang and Geng}{2025}]{chang2025alas}
Edward~Y Chang and Longling Geng.
\newblock {ALAS}: A stateful multi-llm agent framework for disruption-aware planning.
\newblock {\em arXiv preprint arXiv:2505.12501}, 2025.

\bibitem[\protect\citeauthoryear{Chen \bgroup \em et al.\egroup }{2021}]{chen2021evaluating}
Mark Chen, Jerry Tworek, Heewoo Jun, Qiming Yuan, Henrique Ponde De~Oliveira Pinto, Jared Kaplan, Harri Edwards, Yuri Burda, Nicholas Joseph, Greg Brockman, et~al.
\newblock Evaluating large language models trained on code.
\newblock {\em arXiv preprint arXiv:2107.03374}, 2021.

\bibitem[\protect\citeauthoryear{Deng \bgroup \em et al.\egroup }{2025}]{deng2025simura}
Mingkai Deng, Jinyu Hou, Yilin Shen, Hongxia Jin, Graham Neubig, Zhiting Hu, and Eric Xing.
\newblock Simu{RA}: Towards general goal-oriented agent via simulative reasoning architecture with {LLM}-based world model.
\newblock {\em arXiv preprint arXiv:2507.23773}, 2025.

\bibitem[\protect\citeauthoryear{Du \bgroup \em et al.\egroup }{2024}]{du2023improving}
Yilun Du, Shuang Li, Antonio Torralba, Joshua~B Tenenbaum, and Igor Mordatch.
\newblock Improving factuality and reasoning in language models through multiagent debate.
\newblock In {\em Forty-first International Conference on Machine Learning (ICML)}, pages 11733--11763, 2024.

\bibitem[\protect\citeauthoryear{Edge \bgroup \em et al.\egroup }{2024}]{edge2024local}
Darren Edge, Ha~Trinh, Newman Cheng, Joshua Bradley, Alex Chao, Apurva Mody, Steven Truitt, Dasha Metropolitansky, Robert~Osazuwa Ness, and Jonathan Larson.
\newblock From local to global: A graph {RAG} approach to query-focused summarization.
\newblock {\em arXiv preprint arXiv:2404.16130}, 2024.

\bibitem[\protect\citeauthoryear{Erdogan \bgroup \em et al.\egroup }{2025}]{erdogan2025plan}
Lutfi~Eren Erdogan, Nicholas Lee, Sehoon Kim, Suhong Moon, Hiroki Furuta, Gopala Anumanchipalli, Kurt Keutzer, and Amir Gholami.
\newblock Plan-and-act: Improving planning of agents for long-horizon tasks.
\newblock {\em arXiv preprint arXiv:2503.09572}, 2025.

\bibitem[\protect\citeauthoryear{Farquhar \bgroup \em et al.\egroup }{2024}]{farquhar2024detecting}
Sebastian Farquhar, Jannik Kossen, Lorenz Kuhn, and Yarin Gal.
\newblock Detecting hallucinations in large language models using semantic entropy.
\newblock {\em Nature}, 630(8017):625--630, 2024.

\bibitem[\protect\citeauthoryear{Ferrag \bgroup \em et al.\egroup }{2025}]{ferrag2025llm}
Mohamed~Amine Ferrag, Norbert Tihanyi, and Merouane Debbah.
\newblock From {LLM} reasoning to autonomous {AI} agents: A comprehensive review.
\newblock {\em arXiv preprint arXiv:2504.19678}, 2025.

\bibitem[\protect\citeauthoryear{Gero \bgroup \em et al.\egroup }{2023}]{gero2023self}
Zelalem Gero, Chandan Singh, Hao Cheng, Tristan Naumann, Michel Galley, Jianfeng Gao, and Hoifung Poon.
\newblock Self-verification improves few-shot clinical information extraction.
\newblock {\em arXiv preprint arXiv:2306.00024}, 2023.

\bibitem[\protect\citeauthoryear{Guo \bgroup \em et al.\egroup }{2024}]{guo2024large}
Taicheng Guo, Xiuying Chen, Yaqi Wang, Ruidi Chang, Shichao Pei, Nitesh~V Chawla, Olaf Wiest, and Xiangliang Zhang.
\newblock Large language model based multi-agents: A survey of progress and challenges.
\newblock {\em arXiv preprint arXiv:2402.01680}, 2024.

\bibitem[\protect\citeauthoryear{Hariharan \bgroup \em et al.\egroup }{2025}]{hariharan2025plan}
Ananth Hariharan, Vardhan Dongre, Dilek Hakkani-T{\"u}r, and Gokhan Tur.
\newblock Plan verification for {LLM}-based embodied task completion agents.
\newblock {\em arXiv preprint arXiv:2509.02761}, 2025.

\bibitem[\protect\citeauthoryear{He \bgroup \em et al.\egroup }{2024}]{he2024chinese}
Yancheng He, Shilong Li, Jiaheng Liu, Yingshui Tan, Weixun Wang, Hui Huang, Xingyuan Bu, Hangyu Guo, Chengwei Hu, Boren Zheng, et~al.
\newblock Chinese simple{QA}: A chinese factuality evaluation for large language models.
\newblock {\em arXiv preprint arXiv:2411.07140}, 2024.

\bibitem[\protect\citeauthoryear{Hegazy}{2024}]{hegazy2024diversity}
Mahmood Hegazy.
\newblock Diversity of thought elicits stronger reasoning capabilities in multi-agent debate frameworks.
\newblock {\em arXiv preprint arXiv:2410.12853}, 2024.

\bibitem[\protect\citeauthoryear{Huang \bgroup \em et al.\egroup }{2025}]{huang2025deep}
Yuxuan Huang, Yihang Chen, Haozheng Zhang, Kang Li, Huichi Zhou, Meng Fang, Linyi Yang, Xiaoguang Li, Lifeng Shang, Songcen Xu, et~al.
\newblock Deep research agents: A systematic examination and roadmap.
\newblock {\em arXiv preprint arXiv:2506.18096}, 2025.

\bibitem[\protect\citeauthoryear{Jiang \bgroup \em et al.\egroup }{2024}]{jiang2024longrag}
Ziyan Jiang, Xueguang Ma, and Wenhu Chen.
\newblock Long{RAG}: Enhancing retrieval-augmented generation with long-context {LLM}s.
\newblock {\em arXiv preprint arXiv:2406.15319}, 2024.

\bibitem[\protect\citeauthoryear{Kamoi \bgroup \em et al.\egroup }{2024}]{kamoi2024can}
Ryo Kamoi, Yusen Zhang, Nan Zhang, Jiawei Han, and Rui Zhang.
\newblock When can {LLM}s actually correct their own mistakes? a critical survey of self-correction of {LLM}s.
\newblock {\em Transactions of the Association for Computational Linguistics}, 12:1417--1440, 2024.

\bibitem[\protect\citeauthoryear{Kim \bgroup \em et al.\egroup }{2025}]{kim2025reflact}
Jeonghye Kim, Sojeong Rhee, Minbeom Kim, Dohyung Kim, Sangmook Lee, Youngchul Sung, and Kyomin Jung.
\newblock Refl{A}ct: World-grounded decision making in {LLM} agents via goal-state reflection.
\newblock {\em arXiv preprint arXiv:2505.15182}, 2025.

\bibitem[\protect\citeauthoryear{Li \bgroup \em et al.\egroup }{2025}]{li2025websailor}
Kuan Li, Zhongwang Zhang, Huifeng Yin, Liwen Zhang, Litu Ou, Jialong Wu, Wenbiao Yin, Baixuan Li, Zhengwei Tao, Xinyu Wang, et~al.
\newblock Web{S}ailor: Navigating super-human reasoning for web agent.
\newblock {\em arXiv preprint arXiv:2507.02592}, 2025.

\bibitem[\protect\citeauthoryear{Liang \bgroup \em et al.\egroup }{2024}]{liang2024internal}
Xun Liang, Shichao Song, Zifan Zheng, Hanyu Wang, Qingchen Yu, Xunkai Li, Rong-Hua Li, Yi~Wang, Zhonghao Wang, Feiyu Xiong, et~al.
\newblock Internal consistency and self-feedback in large language models: A survey.
\newblock {\em arXiv preprint arXiv:2407.14507}, 2024.

\bibitem[\protect\citeauthoryear{Luo \bgroup \em et al.\egroup }{2025}]{luo2025large}
Junyu Luo, Weizhi Zhang, Ye~Yuan, Yusheng Zhao, Junwei Yang, Yiyang Gu, Bohan Wu, Binqi Chen, Ziyue Qiao, Qingqing Long, et~al.
\newblock Large language model agent: A survey on methodology, applications and challenges.
\newblock {\em arXiv preprint arXiv:2503.21460}, 2025.

\bibitem[\protect\citeauthoryear{Manakul \bgroup \em et al.\egroup }{2023}]{2023selfcheckgpt}
Potsawee Manakul, Adian Liusie, and Mark~JF Gales.
\newblock {SELFCHECKGPT}: Zero-resource black-box hallucination detection for generative large language models.
\newblock {\em arXiv preprint arXiv:2303.08896}, 2023.

\bibitem[\protect\citeauthoryear{Mialon \bgroup \em et al.\egroup }{2024}]{mialon2023gaia}
Gr{\'e}goire Mialon, Cl{\'e}mentine Fourrier, Thomas Wolf, Yann LeCun, and Thomas Scialom.
\newblock {GAIA}: a benchmark for general {AI} assistants.
\newblock In {\em The Twelfth International Conference on Learning Representations (ICLR)}, 2024.

\bibitem[\protect\citeauthoryear{Muhammed \bgroup \em et al.\egroup }{2025}]{muhammed2025selfcheckagent}
Diyana Muhammed, Gollam Rabby, and S{\"o}ren Auer.
\newblock Self{C}heck{A}gent: Zero-resource hallucination detection in generative large language models.
\newblock {\em arXiv preprint arXiv:2502.01812}, 2025.

\bibitem[\protect\citeauthoryear{Pan \bgroup \em et al.\egroup }{2023}]{pan2024autocorrect}
Liangming Pan, Michael Saxon, Wenda Xu, Deepak Nathani, Xinyi Wang, and William~Yang Wang.
\newblock Automatically correcting large language models: Surveying the landscape of diverse self-correction strategies.
\newblock {\em arXiv preprint arXiv:2308.03188}, 2023.

\bibitem[\protect\citeauthoryear{Parmar \bgroup \em et al.\egroup }{2025}]{parmar2025plangen}
Mihir Parmar, Xin Liu, Palash Goyal, Yanfei Chen, Long Le, Swaroop Mishra, Hossein Mobahi, Jindong Gu, Zifeng Wang, Hootan Nakhost, et~al.
\newblock {PlanGEN}: A multi-agent framework for generating planning and reasoning trajectories for complex problem solving.
\newblock {\em arXiv preprint arXiv:2502.16111}, 2025.

\bibitem[\protect\citeauthoryear{Phan \bgroup \em et al.\egroup }{2025}]{phan2025humanity}
Long Phan, Alice Gatti, Ziwen Han, Nathaniel Li, Josephina Hu, Hugh Zhang, Chen Bo~Calvin Zhang, Mohamed Shaaban, John Ling, Sean Shi, et~al.
\newblock Humanity's last exam.
\newblock {\em arXiv preprint arXiv:2501.14249}, 2025.

\bibitem[\protect\citeauthoryear{Plaat \bgroup \em et al.\egroup }{2024}]{plaat2024reasoning}
Aske Plaat, Annie Wong, Suzan Verberne, Joost Broekens, Niki van Stein, and Thomas B{\"a}ck.
\newblock Reasoning with large language models, a survey.
\newblock {\em Computing Research Repository (CoRR)}, 2024.

\bibitem[\protect\citeauthoryear{Qiao \bgroup \em et al.\egroup }{2025}]{qiao2025webresearcher}
Zile Qiao, Guoxin Chen, Xuanzhong Chen, Donglei Yu, Wenbiao Yin, Xinyu Wang, Zhen Zhang, Baixuan Li, Huifeng Yin, Kuan Li, et~al.
\newblock Web{R}esearcher: Unleashing unbounded reasoning capability in long-horizon agents.
\newblock {\em arXiv preprint arXiv:2509.13309}, 2025.

\bibitem[\protect\citeauthoryear{Shinn \bgroup \em et al.\egroup }{2023}]{shinn2023reflexion}
Noah Shinn, Federico Cassano, Ashwin Gopinath, Karthik Narasimhan, and Shunyu Yao.
\newblock Reflexion: Language agents with verbal reinforcement learning.
\newblock {\em Advances in Neural Information Processing Systems (NeurIPS)}, 36:8634--8652, 2023.

\bibitem[\protect\citeauthoryear{Tang \bgroup \em et al.\egroup }{2024}]{tang2024minicheck}
Liyan Tang, Philippe Laban, and Greg Durrett.
\newblock Mini{C}heck: Efficient fact-checking of {LLM}s on grounding documents.
\newblock {\em arXiv preprint arXiv:2404.10774}, 2024.

\bibitem[\protect\citeauthoryear{Verma \bgroup \em et al.\egroup }{2024}]{verma2024brittle}
Mudit Verma, Siddhant Bhambri, and Subbarao Kambhampati.
\newblock On the brittle foundations of {ReAct} prompting for agentic large language models.
\newblock {\em arXiv preprint arXiv:2405.13966}, 2024.

\bibitem[\protect\citeauthoryear{Wang \bgroup \em et al.\egroup }{2022}]{wang2022self}
Xuezhi Wang, Jason Wei, Dale Schuurmans, Quoc Le, Ed~Chi, Sharan Narang, Aakanksha Chowdhery, and Denny Zhou.
\newblock Self-consistency improves chain of thought reasoning in language models.
\newblock {\em arXiv preprint arXiv:2203.11171}, 2022.

\bibitem[\protect\citeauthoryear{Yao \bgroup \em et al.\egroup }{2023a}]{yao2023tree}
Shunyu Yao, Dian Yu, Jeffrey Zhao, Izhak Shafran, Tom Griffiths, Yuan Cao, and Karthik Narasimhan.
\newblock Tree of thoughts: Deliberate problem solving with large language models.
\newblock {\em Advances in neural information processing systems (NeurIPS)}, 36:11809--11822, 2023.

\bibitem[\protect\citeauthoryear{Yao \bgroup \em et al.\egroup }{2023b}]{yao2023react}
Shunyu Yao, Jeffrey Zhao, Dian Yu, Nan Du, Izhak Shafran, Karthik Narasimhan, and Yuan Cao.
\newblock Re{A}ct: Synergizing reasoning and acting in language models.
\newblock In {\em International Conference on Learning Representations (ICLR)}, 2023.

\bibitem[\protect\citeauthoryear{Yu \bgroup \em et al.\egroup }{2025a}]{yu2025aworld}
Chengyue Yu, Siyuan Lu, Chenyi Zhuang, Dong Wang, Qintong Wu, Zongyue Li, Runsheng Gan, Chunfeng Wang, Siqi Hou, Gaochi Huang, et~al.
\newblock {AWorld}: Orchestrating the training recipe for agentic {AI}.
\newblock {\em arXiv preprint arXiv:2508.20404}, 2025.

\bibitem[\protect\citeauthoryear{Yu \bgroup \em et al.\egroup }{2025b}]{yu2025dyna}
Xiao Yu, Baolin Peng, Ruize Xu, Michel Galley, Hao Cheng, Suman Nath, Jianfeng Gao, and Zhou Yu.
\newblock Dyna-{T}hink: Synergizing reasoning, acting, and world model simulation in ai agents.
\newblock {\em arXiv preprint arXiv:2506.00320}, 2025.

\bibitem[\protect\citeauthoryear{Zhang \bgroup \em et al.\egroup }{2025a}]{zhang2025agentorchestra}
Wentao Zhang, Liang Zeng, Yuzhen Xiao, Yongcong Li, Ce~Cui, Yilei Zhao, Rui Hu, Yang Liu, Yahui Zhou, and Bo~An.
\newblock {AgentOrchestra}: A hierarchical multi-agent framework for general-purpose task solving.
\newblock {\em arXiv preprint arXiv:2506.12508}, 2025.

\bibitem[\protect\citeauthoryear{Zhang \bgroup \em et al.\egroup }{2025b}]{zhang2025survey}
Zeyu Zhang, Quanyu Dai, Xiaohe Bo, Chen Ma, Rui Li, Xu~Chen, Jieming Zhu, Zhenhua Dong, and Ji-Rong Wen.
\newblock A survey on the memory mechanism of large language model-based agents.
\newblock {\em ACM Transactions on Information Systems}, 43(6):1--47, 2025.

\end{thebibliography}

\end{document}